\documentclass[sigconf]{acmart}

\usepackage{graphicx} 
\usepackage{color}
\usepackage[english]{babel}
\usepackage[utf8]{inputenc}
\usepackage{algorithm}
\usepackage[noend]{algpseudocode}
\usepackage{multirow}
\usepackage{multicol}
\usepackage{enumitem}
\setlist[itemize]{leftmargin=*}
\usepackage{float}

\AtBeginDocument{%
  \providecommand\BibTeX{{%
    \normalfont B\kern-0.5em{\scshape i\kern-0.25em b}\kern-0.8em\TeX}}}

\copyrightyear{2024}
\acmYear{2024}
\setcopyright{acmlicensed}
\acmConference[WSDM '24]{Proceedings of the 17th ACM International Conference on Web Search and Data Mining}{March 4--8, 2024}{Merida, Mexico}
\acmBooktitle{Proceedings of the 17th ACM International Conference on Web Search and Data Mining (WSDM '24), March 4--8, 2024, Merida, Mexico}
\acmPrice{15.00}
\acmDOI{10.1145/3616855.3635793}
\acmISBN{979-8-4007-0371-3/24/03}

\begin{document}

\title{Distribution Consistency based Self-Training for Graph Neural Networks with Sparse Labels}

\author{Fali Wang}
\affiliation{%
  \institution{The Pennsylvania State University}
  \city{University Park}
  \country{USA}}
\email{fqw5095@psu.edu}

\author{Tianxiang Zhao}
\affiliation{%
  \institution{The Pennsylvania State University}
  \city{University Park}
  \country{USA}}
\email{tkz5084@psu.edu}

\author{Suhang Wang}
\affiliation{%
  \institution{The Pennsylvania State University}
  \city{University Park}
  \country{USA}}
\email{szw494@psu.edu}


\begin{abstract}
Few-shot node classification poses a significant challenge for Graph Neural Networks (GNNs) due to insufficient supervision and potential distribution shifts between labeled and unlabeled nodes. Self-training has emerged as a widely popular framework to leverage the abundance of unlabeled data, which expands the training set by assigning pseudo-labels to selected unlabeled nodes. Efforts have been made to develop various selection strategies based on confidence, information gain, etc. However, none of these methods takes into account the distribution shift between the training and testing node sets. The pseudo-labeling step may amplify this shift and even introduce new ones, hindering the effectiveness of self-training. Therefore, in this work, we explore the potential of explicitly bridging the distribution shift between the expanded training set and test set during self-training. 
To this end, we propose a novel Distribution-Consistent Graph Self-Training (DC-GST) framework to identify pseudo-labeled nodes that both are informative and capable of redeeming the distribution discrepancy and formulate it as a differentiable optimization task. A distribution-shift-aware edge predictor is further adopted to augment the graph and increase the model's generalizability in assigning pseudo labels.
We evaluate our proposed method on four publicly available benchmark datasets and extensive experiments demonstrate that our framework consistently outperforms state-of-the-art baselines.

\end{abstract}

%
%



\maketitle




\section{Introduction}
Graph Neural Networks (GNNs) have achieved noteworthy success on semi-supervised
node classification on graphs, which exploit the graph inductive bias and labeled nodes to predict labels for the remaining unlabeled nodes \cite{kipf2017semisupervised,velivckovicgraph,hamilton2017inductive,bo2021beyond,abu2019mixhop,MAURYA2022101695}. However, their advance is often dependent on intricate training with a large number of labeled nodes, which can be costly to acquire \cite{sun2020multi}. Empirically, the performance of Graph Convolutional Networks (GCNs) tends to deteriorate rapidly as the amount of labeled data decreases \cite{zhou2019effective}. The generalization capacity of the classifier is significantly affected by the variability of the labeled nodes \cite{zhu2021shift}. Biases in the sampling process can result in distributional shifts
between the training nodes and the rest of the graph, causing the classifier to over-fit in the training data irregularities and resulting in performance degradation during testing. For example, in social networks, popular users (high-degree nodes) are more likely to be labeled. However, as social networks follow a power-law distribution, the majority of nodes are low-degree nodes. Those popular nodes are actually not representative nodes, causing a distribution shift in terms of both graph structure and node attributes.  Furthermore, it is usually costly and expensive to obtain labels. Thus, we are often faced with sparse labels. 
The sparse label also makes the distribution shifts more severe due to the less coverage of representative labeled nodes.

\begin{figure}[!t]
    \vskip -1.5em
    \centering
    \includegraphics[width=0.24\textwidth]{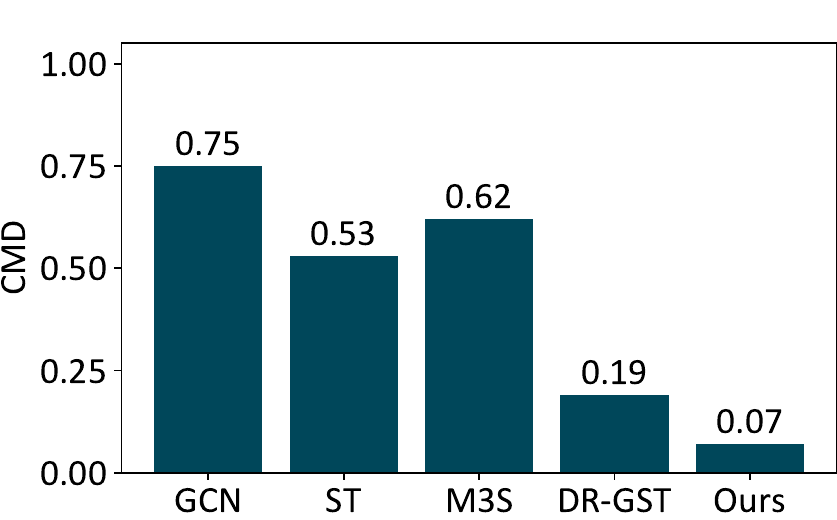}
    \vskip -1em
    \caption{CMD on Cora: minimal distribution shift with the proposed method using GCN backbone against advanced self-training methods including ST, M3S, and DR-GST. Note that the value from GCN signifies the initial shift, as the CMD computation is from the high-level embedding space. }
    \label{fig:fig1}
    \vskip -1.5em
\end{figure}

Self-training \cite{li2018deeper,lee2013pseudo} has shown great success in tackling sparse labels. 
Generally, in self-training, an arbitrary backbone trained on the original labeled data acts as the \textbf{teacher} model. This teacher model assigns pseudo-labels to selected unlabeled nodes and incorporates them to augment the labeled data, after which a \textbf{student} model is trained on the augmented training nodes. Typically, current self-training frameworks introduce multiple stages where the teacher model is iteratively updated with augmented data to generate more confident pseudo-labeled nodes \cite{sun2020multi,zhou2019effective,9703190,ding2022learning}. For example, Self-Training (ST)~\cite{li2018deeper} represents a single-stage self-training framework that initially trains a GCN using given labels, subsequently selects predictions exhibiting the highest confidence, and then proceeds to further train the GCN utilizing the expanded set of labels. Conversely, M3S~\cite{sun2020multi} stands for a multi-stage self-training framework that, during each stage, applies DeepCluster to the graph embeddings of GCNs and introduces an alignment mechanism for the clusters to generate pseudo-labels.
Despite their great success, most existing self-training methods are developed under the assumption of distribution consistency across labeled and unlabeled node sets. Yet, a distribution shift often occurs between these two sets. This distribution shift is more severe when the labeled nodes are sparse because there are not enough labeled nodes to well represent the distribution. In addition, recent work~\cite{liu2022confidence} shows that the pseudo-labeling step may amplify this shift and even introduce new shifts, which are overlooked by these methods. One related work is DR-GST \cite{liu2022confidence}, which is proposed to alleviate new shifts brought by the pseudo-labeling process. However, it fails to address the distribution shift that already presents with train-test splits. To verify this, we craft a training set on Cora with distribution shift following~\cite{zhu2021shift}, which enables us to evaluate the performance of existing methods under distribution shifts with sparse training labels. Central Moment Discrepancy (CMD) metric \cite{zellinger2019robust} is used to measure the distribution shifts of learned node representations between labeled nodes and test nodes. The larger CMD is, the more severe the distribution shift is. Fig. \ref{fig:fig1} shows the results. From the figure, we can observe that the persistence of distribution shift is evident when comparing the backbone to self-training methods, such as ST and M3S, that do not incorporate considerations for distribution shift mitigation.
Although DR-GST can account for distribution shifts in pseudo-labeling, it fails to eliminate shifts present in the original data. Consequently, biased pseudo-labeled nodes hinder the GNN performance and result in more biased pseudo-labels during subsequent pseudo-labeling stages, making it difficult to leverage the benefits of multiple-stage self-training. 
Hence, it is important to address the distribution shift between the initial sparse labels and the abundant unlabeled nodes and the distribution shift introduced during pseudo-labeling. However, the work in this direction is rather limited.  


\begin{figure}[!t]
    \centering
    \includegraphics[width=0.4\textwidth]{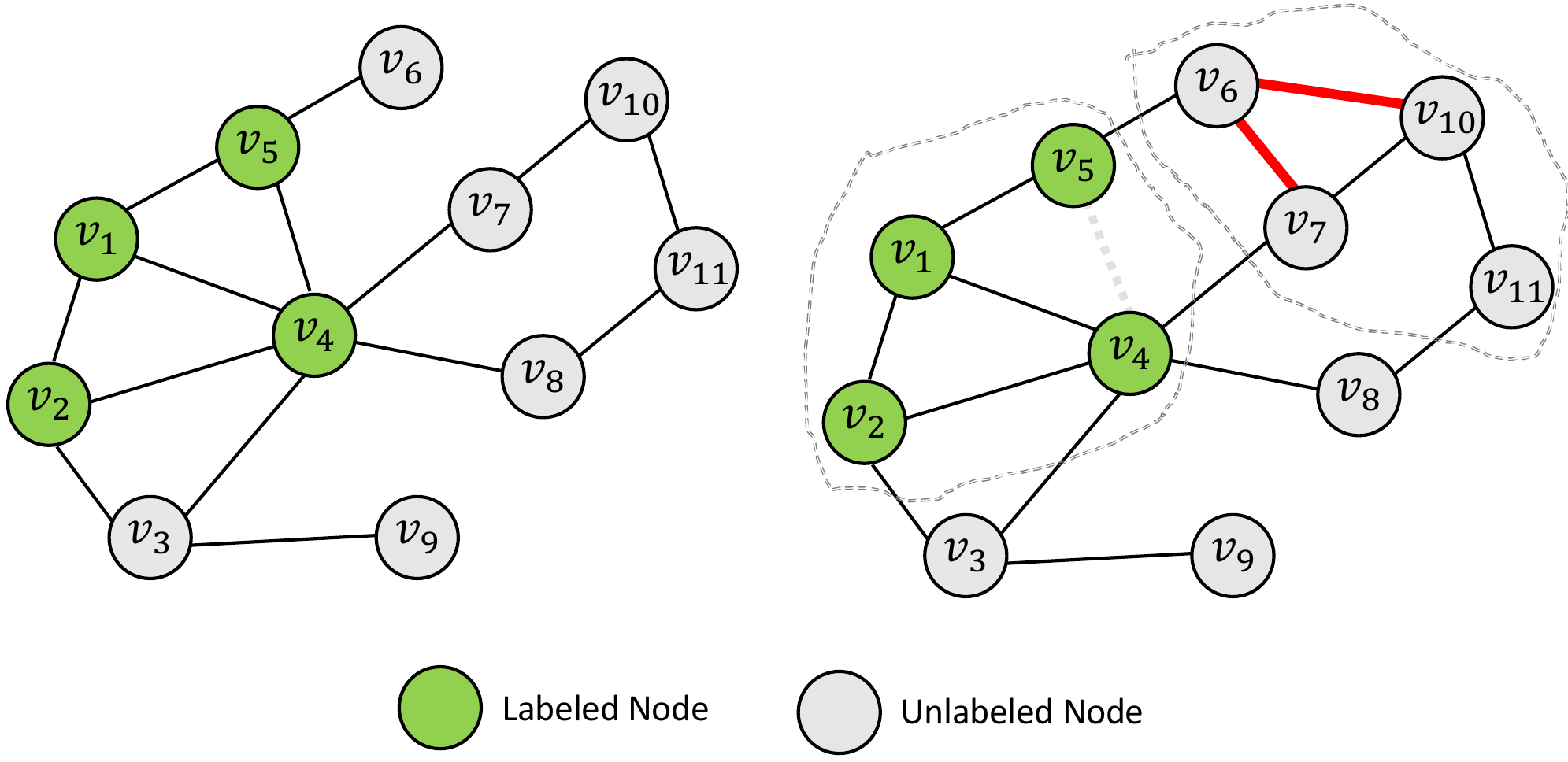}
    \vskip -1.5em
    \caption{The comparison between original and variant graphs. Higher consistency (e.g., in circled areas) can be achieved with EP(CMD).}
    \label{fig:toy_example1}
    \vskip -1.5em
\end{figure}

Therefore, in this paper, we investigate a new problem setting of few-shot node classification under distribution shift during self-training.
In essence, there are two challenges: (i) data sparsity; (ii) distribution shifts. To address these challenges,  we propose a novel \textbf{D}istribution \textbf{C}onsistency based \textbf{G}raph \textbf{S}elf-\textbf{T}raining (DC-GST) framework which can utilize the self-training paradigm to mitigate the former one and reduce the distribution distance in GNNs to overcome the latter one.  
Drawing inspiration from existing work on shift robustness \cite{zhu2021shift}, we introduce a \textbf{Distribution Consistency (DC)} criterion which identifies nodes capable of compensating for the current distribution shift. This criterion not only mitigates the distributional discrepancy engendered by pseudo-labeling but also lessens the shift within the training set. It focuses on selecting pseudo-labeled nodes that minimize the distribution distance between the augmented training set and the test set, i.e. diminishing the CMD distance. The augmented training set consists of the ground-truth labeled nodes and a subset of candidate nodes, where the selection of the latter one is optimized to minimize the distance. 
Nevertheless, owing to the sparsity of annotated data and the accompanying distributional shift issue, obtaining reliable pseudo-labels is tough. To tackle it, we augment the original graph with an \textbf{Edge Predictor (EP)} inspired by 
the evidence in~\cite{Zhao_Liu_Neves_Woodford_Jiang_Shah_2021} that the neural edge predictor can implicitly learn class-homophilic tendencies in existing edges deemed improbable and in absent edges considered probable. Revising edges is a simple yet effective strategy to augment the original graph, particularly under weak supervision scenarios~\cite{Zhao_Liu_Neves_Woodford_Jiang_Shah_2021}. It generates a graph variant by deterministically adding (removing) new (existing) edges to increase the graph's diversity. EP augments the original graph structures under two objectives: creating more diverse inputs and reducing the distribution shifts across training and test nodes. We name the edge predictor with reduced CMD metric as \textbf{EP(CMD)} module. Once we equip our teacher model with the EP, it gains exposure to a broad variety of graphs, as illustrated in Fig. \ref{fig:toy_example1}, which mitigates training-testing shifts. Furthermore, to select informative pseudo-labeled nodes, we use a \textbf{Neighborhood Entropy Reduction (NER)} criterion in addition to the aforementioned DC in order to maximize knowledge that selected pseudo labels can offer to the neighborhood. The underlying intuition is that a node is most informative when it can maximally reduce the uncertainty (i.e. entropy) within its neighborhood. However, obtaining a closed-form solution for selection criteria poses a challenge due to the exponential search space and the non-convex optimization objective. Heuristic greedy search techniques may fall short due to the interdependence among nodes. To circumvent these obstacles, we transform the selection task into a differentiable optimization problem. This reformulation allows the application of gradient descent for problem-solving. 
In summary, our main contributions are as follows: 
\begin{itemize}[leftmargin=*]
    \item We identify and study an important new problem, addressing the challenge of distribution shift in graph self-training. 
    \item We propose a novel framework, DC-GST, which incorporates a distribution consistency-based criterion and a neighborhood entropy reduction criterion for pseudo-labeled nodes. This framework also leverages a specialized edge predictor module to enhance the generalization of the teacher model.
    \item Experimental results show the effectiveness of the proposed DC-GST in few shot node classification with distribution shift.
\end{itemize}
\section{Related Work}

\subsection{Graph Neural Networks}
Graph Neural Networks (GNNs) have become a focus in recent years, comprising two primary types: spectral-based and spatial-based GNNs. Spectral-based GNNs, exemplified by the Graph Convolutional Network (GCN) \cite{kipf2017semisupervised}, leverage graph convolution as a fundamental operation within the spectral domain. 
On the other hand, spatial-based GNNs, such as the Graph Attention Network (GAT) \cite{velivckovicgraph} and the GSAGE \cite{hamilton2017inductive}, emphasize the spatial domain, focusing on local information aggregation and transformation. 
Despite their widespread adoption, GNNs, including GCN, GAT, and GSAGE, possess inherent limitations, particularly when confronted with sparse label scenarios. This stems from the localized nature of the convolutional filter or the aggregation mechanism, which hampers the effective propagation of labels across the entire graph when only a few labels are available \cite{li2018deeper}.
Moreover, the potential bias in the sampling process, utilized to select nodes for training, can engender distribution discrepancies between the training set and the remaining graph, potentially leading to overfitting during inference due to the uneven labeling \cite{zhu2021shift}. Even contemporary benchmarks for GNNs assume the availability of independent and identically distributed (IID) samples for training labels \cite{hu2020open}. Remarkably, the problem of distribution shift has been ignored in most work on semi-supervised learning using GNNs for node classification tasks, albeit its potential implications. Therefore, addressing the issue of distribution shift is pivotal for GNN advancements. 


\subsection{Graph Self-Training} 

To combat label scarcity, self-training \cite{scudder1965probability,zoph2020rethinking} has emerged as a promising approach to leverage abundant unlabeled data, delivering remarkable performance across several tasks \cite{lee2013pseudo,mukherjee2020uncertainty,rosenberg2005semi}. In the graph domain, graph self-training has been deployed to overcome the limitation of propagating labels to the entire graph in a shallow GNN \cite{li2018deeper}.
At the heart of the success of self-training lies the pseudo-labeling strategy, essentially the method for selecting appropriate nodes to augment the training set. Confidence-based self-training is one such strategy, prevalent in existing self-training frameworks \cite{sun2020multi,zhou2019effective,li2018deeper,wang2021confident,zoph2020rethinking}, that selects high-confidence unlabeled nodes for training set expansion and subsequent model retraining.
MT-GCN \cite{zhan2021mutual} further improved self-training performance by training dual GCN models through mutual teaching, using the highest-confidence pseudo labels for learning. M3S \cite{sun2020multi} incorporated the deep cluster and alignment mechanism into the multi-stage training framework, while DSGSN \cite{zhou2019effective} introduced a negative sampling regularization technique to address the unstable training issue of existing GNNs. \cite{pedronette2021rank} used a margin score through a rank-based model to identify the most confident sample predictions. However, those nodes with the most confident pseudo labels have been deemed as less-informative and useless \cite{liu2022confidence}, as most of them lie far from the decision boundary. 
To address it, recent studies have presented information-based pseudo-labeling as an alternative solution. DR-GST \cite{liu2022confidence} proposes to employ information gain to reweight each node during retraining, thereby mitigating the effects of useless pseudo labels. An informativeness module is introduced to learn via contrastive learning to select pseudo labels with high information~\cite{li2023informative}. 
Yet, none of these methods have considered the innate distribution shift between the training and test set.

\subsection{Distribution Shifts on Graphs} 
In most node classification benchmarks, it is assumed that the training and testing node samples have an identical distribution. However, in real-world scenarios, distribution shifts, such as covariate shifts, frequently occur between the labeled training set and the unlabeled testing set~\cite{wu2022recent}. As a result, a GNN classifier may overfit to the irregularities of the training data, adversely affecting its performance post-deployment. Domain Invariant Representation (DIR) Learning seeks to learn generalizable node embeddings by minimizing their distributional discrepancy across various domains, a process that can be quantified through adversarial learning \cite{ganin2016domain} or heuristic metrics such as CMD \cite{zellinger2019robust} and MMD \cite{long2017deep}. The primary aim of these methodologies is to bridge the gap caused by distribution shifts across different domains, applicable in both inductive \cite{bevilacqua2021size} and transductive settings \cite{zhu2021shift}. Further studies have explored the development of more generalizable or robust GNNs by augmenting topology structures either during training~\cite{rong2019dropedge,luo2021learning} or during inference~\cite{brody2021attentive,dai2022comprehensive}. In this study, we pioneer the exploration of distribution shifts in graphs within the context of self-training.


\section{Preliminary} 

\subsection{Notations and GNNs} 

We use $\mathcal{G}=\{ \mathcal{V}, \mathcal{E}, \mathbf{X} \}$ to denote an attributed graph, where $\mathcal{V}=\{v_1, v_2, \dots, v_n\}$ is the set of $n$ nodes and $\mathcal{E}$ is the set of edges. $\mathbf{X} \in \mathcal{R}^{|\mathcal{V}| \times D_v}$ is the node attribute matrix, where $\mathbf{x}_i$ is the $D_v$ dimensional attribute vector of $v_i$. The edges can be described by an adjacency matrix $\mathbf{A}$, where ${A}_{ij}=1$ if node $v_i$ and $v_j$ are connected; otherwise ${A}_{ij}=0$. 
In semi-supervised node classification, the labels of a small portion of nodes are known. We use $\mathcal{V}_L$ to denote the labeled node set and $\mathbf{Y}_L$ to denote the corresponding labels. The goal is to leverage both $\mathcal{G}$ and $\mathbf{Y}_L$ to predict the labels of unlabeled nodes. 

Graph Neural Networks (GNNs) have achieved great success in semi-supervised node classification on graphs. Generally, GNN layers can be summarized as a message-passing framework, which iteratively propagates and aggregates messages in the neighborhood.
Using $\mathbf{h}_v^l$ to denote the representation of node $v$ at the $l$-th GNN layer with $\mathbf{h}_v^0 := \mathbf{x}_v$, the inference process can be written as:
    \begin{equation}
        \begin{aligned}
        \mathbf{m}_{v}^{l+1} &=\sum_{u \in \mathcal{N}(v)} \text{Message}^{l}\large( \mathbf{h}_v^l, \mathbf{h}_u^l, A_{v,u} \large), \\
        \mathbf{h}_v^{l+1} &= \text{Update}^{l}\large(\mathbf{h}_v^{l}, \mathbf{m}_{v}^{l+1} \large),
        \end{aligned}
    \end{equation}
where $\text{Message}^{l}$ and $\text{Update}^{l}$ are the message function and update function at $l$-th layer, respectively. $\mathcal{N}(v)$ is the set of node $v$'s neighbors. Taking a GCN layer for instance, the message function is parameterized by a to-be-learned linear layer and is re-weighted based on the normalized Laplacian, while the update function is implemented as the summation. Formally, a GCN layer can be written into the following equation:
\begin{equation}
\mathbf{H}^{l+1} = \phi(\tilde{\mathbf{A}} \mathbf{H}^{l} \theta^k)
\end{equation}
where $\tilde{\mathbf{A}}$ is a normalized adjacency matrix $\mathbf{D}^{-\frac{1}{2}}(\mathbf{A}+\mathbf{I})\mathbf{D}^{-\frac{1}{2}}$, $\mathbf{D}$ is the degree matrix of $(\mathbf{A}+\mathbf{I})$, $\mathbf{I}$ is the identity matrix, $\phi$ is the activation function. We refer to the final latent representations as $\mathbf{Z}$ throughout the work for brevity. There exist multiple variants of GNN layers, such as GCN \cite{kipf2017semisupervised}, GAT \cite{velivckovicgraph}, and GSAGE \cite{hamilton2017inductive}. 

\subsection{Graph Self-Training}

Graph Self-Training (GST), an established methodology for bolstering Graph Neural Networks (GNNs) performance, strategically leverages additional pseudo labels to address the insufficiency of labeled data. The method initiates by training a GNN on the limited supply of available labels, following which the GNN generates predictions for unlabeled nodes and collects the most reliable pseudo labels to complement the original label set. This augmented set is subsequently employed to retrain another de novo GNN (with the same model structure) \cite{li2018deeper} which is utilized to produce predictions for unlabeled nodes.
Recent advancements \cite{liu2022confidence,sun2020multi} have evolved this paradigm into a multi-stage self-training scheme, characterized by multiple iterations of training the teacher GNN model, in each stage employing the pseudo labels of the preceding stage. In the initial stage, the teacher GNN is trained on a ground-truth labeled training set. As we move to the subsequent stages, the teacher model is trained on an augmented label set, which incorporates both the original labels and the most reliable pseudo-labels from the previous stage. At each stage, the most reliable pseudo labels (who often carry the most confidence or richest information)  are merged into the augmented label set for the following stage.
The final step of this process is to train the student GNN model on the augmented label set of the final stage so that the student model can produce accurate predictions for unlabeled nodes.


\subsection{Distribution Shift and Problem Definition}
The objective of graph out-of-distribution generalization is to develop an optimal graph predictor which is capable of achieving superior generalization on data points originating from the test distribution $P_{test}(x, y)$, given the training set derived from a distinct training distribution $P_{train}(x,y)$, where $P_{train}(x,y) \neq P_{test}(x,y)$. The distributional shift between $P_{train}(x,y)$ and $P_{test}(x,y)$ could lead to the breakdown of a graph predictor based on the IID hypothesis since direct minimization of the average loss over training instances fails to generate an optimal predictor capable of generalizing to testing instances under distribution shifts \cite{li2022out}.

Defining distribution shifts on a graph is challenging due to the difficulty of measuring graph structure information. As a solution, \cite{zhu2021shift} propose a definition in the representation space of a GNN.

\begin{definition}[Distribution shift in GNNs] Assume node representations $\mathbf{Z} = \{\mathbf{z}_1, \mathbf{z}_2, . . . , \mathbf{z}_{|\mathcal{V}|} \}$ are given as an output of the last hidden layer of a Graph Neural Network (GNN) on a graph $\mathcal{G}$. Given labeled data ${(\mathbf{x}_i, y_i)}$ of size $|\mathcal{V}_L|$, the labeled node representation, $\mathbf{Z}_L = \{ \mathbf{z}_1, . . . , \mathbf{z}_{|\mathcal{V}_L|} \}$, is a subset of the node representations. The distribution shift in GNNs is then measured via a distance metric $D(\mathbf{Z}, \mathbf{Z}_L)$, such as Maximum Mean Discrepancy (MMD) or Central Moment Discrepancy (CMD).
\end{definition}

In the few-shot setting,  the scarcity of labeled nodes critically limits their representativeness, thereby exacerbating the issue of distribution shift. Furthermore, even a confidence-based self-training strategy tends to maintain or widen this shift, as the pseudo-labeled instances are concentrated around nodes with ground-truth labels to a large extent. Thus, we aim to investigate the distribution shift during self-training specifically within few-shot scenarios. The problem is formally defined as:

\vspace{1em}

\noindent\textbf{Problem Definition.} Given a graph $\{\mathcal{V}, \mathcal{E}, \mathbf{X}\}$ and a limited labeled nodes $\mathcal{V}_L$ with labels $\mathbf{Y}_L$, there exists distribution shifts between labeled node set $\mathcal{V}_L$ and unlabeled node set. 
Our goal is to devise a graph self-training framework utilizing a GNN node classifier $f_\theta$ to predict $\hat{y}_u = \arg\max_j f_\theta(v_u)_j$ for every unlabeled node $v_u \in \mathcal{V}_U$. 


\begin{figure}[!t]
    \centering
    \includegraphics[width=0.5\textwidth]{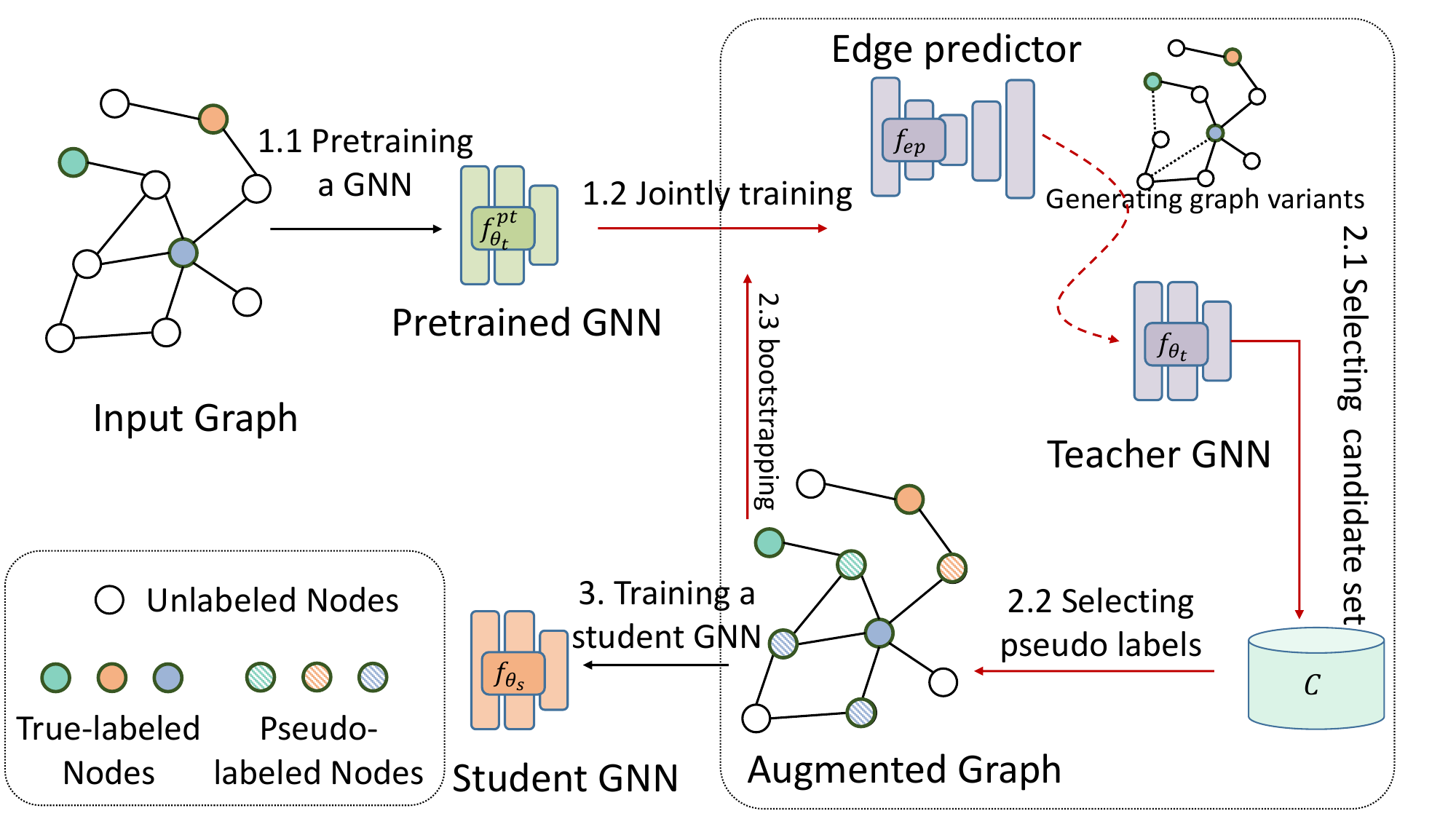}
    \vskip -1.5em
    \caption{The proposed distribution consistency based graph self-training framework. Red arrows indicate the loop.}
    \label{fig:framework}
    \vspace{-1em}
\end{figure}

\section{Proposed Method}

In this section, we elaborate on a novel framework Distribution Consistency based Graph Self-Training (DC-GST). The framework involves two stages: self-training a teacher GNN with updated pseudo-labeled nodes and training a student GNN with the final pseudo labels, as shown in Fig. \ref{fig:framework}. During the self-training process, the teacher GNN is iteratively trained using a combination of ground-truth labeled nodes and selected pseudo-labeled nodes, and then assigns pseudo labels to the unlabeled nodes. The selection of pseudo-labeled nodes is based on the distribution consistency and neighborhood entropy reduction criteria. After the iterative update of the teacher GNN, the final step is employed to train a student GNN by incorporating an augmented training set, obtaining the final prediction. We discuss the details of the DC-GST framework in the following section.




\subsection{Generalized Teacher GNN with Few Labels} 
To overcome the limitation of graph convolution's localized nature when few labeled nodes are available, iterative self-training is employed. The first step of each iteration is to train the teacher GNN on the (pseudo-)labeled data $\mathcal{C}_A$. During the initial iteration, $\mathcal{C}_A$ is set to $\mathcal{V}_L$. The goal is to assign pseudo labels to all unlabeled nodes. However, the label scarcity issue, particularly in the first few stages, and the distribution shift issue, stemming from both the training set and pseudo-labeling, can lead to the teacher GNN overfitting to the few-shot labeled node distribution.
In turn, this can render the generated pseudo-labels unreliable, hampering the effectiveness of pseudo-labels. Therefore, training a generalized teacher GNN is essential. We mitigate the former issue by implementing graph data augmentation. This approach rectifies the model's bias towards few-shot labeled node distribution by exposing it to more graph variants. 
To address the latter issue of distribution shift, we entail the selection of graph variants that exhibit the minimal distributional distance between the (pseudo-)labeled data and test nodes. Edge prediction serves as a mechanism to align the structural distribution of labeled and unlabeled nodes, thereby enhancing message passing from labeled to unlabeled nodes.

Concretely, we introduce an innovative graph data augmentation technique to bolster the generalization of the teacher GNN. This technique revises graph structures with an \textbf{Edge Predictor (EP)}, denoted as $f_{ep}$: $\mathbf{A},\mathbf{X} \rightarrow \mathbf{A}'$, which can implicitly learn class-homophilic tendencies in existing edges deemed improbable and in absent edges considered probable~\cite{Zhao_Liu_Neves_Woodford_Jiang_Shah_2021}. Revising edges is a simple yet effective strategy to augment the original graph, particularly under weak supervision scenarios~\cite{Zhao_Liu_Neves_Woodford_Jiang_Shah_2021}. It generates a graph variant by deterministically adding (removing) new (existing) edges to increase the graph's diversity. Specifically, the EP module 
first generates an edge probability matrix with the same shape as the adjacency matrix as
\begin{equation}
    \mathbf{M} = \phi(\mathbf{T}\mathbf{T}^\top), \text{ where }  \mathbf{T}=f_{GCN}(\mathbf{A}, \mathbf{X})
\end{equation}
Here $f_{GCN}$ is a two-layer GCN to learn node representation. Next, in the edge sampling phase, the matrix is sparsified using relaxed Bernoulli sampling to generate the graph variant adjacency matrix $\mathbf{A}'$. This sampling approach is a binary special case of the Gumbel-Softmax reparameterization trick \cite{maddison2017the}: 
\begin{align}
    \begin{aligned}
        \mathbf{A}'_{ij} = \left \lfloor \frac{1}{1+\exp(-(\log \mathbf{M}_{ij}+G)/\tau)} + \frac{1}{2} \right \rfloor
        \label{eq:edge_predictor}
    \end{aligned}
\end{align}
where $\tau$ is the temperature of Gumbel-Softmax distribution, and $G \sim$ Gumbel(0, 1) is a Gumbel random variate. We specify $\tau=1.2$ as recommended in \cite{Zhao_Liu_Neves_Woodford_Jiang_Shah_2021}. The adjacency matrix $\mathbf{A}'$ and feature matrix $\mathbf{X}$ are fed into the teacher GNN to get node representations as
\begin{equation} \label{eq:Z}
    \mathbf{Z}=f_{\theta_t}(\mathbf{A}', \mathbf{X})
\end{equation}

Moreover, we want the teacher model to derive an embedding space that is distributionally consistent across training and test nodes. To accomplish this, we integrate a distributionally consistent objective on node representations $\mathbf{Z}$ from the graph variant. 
This strategy can guarantee that the teacher model is exposed to more graph variants with consistent distributions between training and test nodes. We minimize the distribution distance between the training and test set explicitly on the EP module. 
The quantification of shifts can be effectively achieved through the use of discrepancy metrics such as Central Moment Discrepancy (CMD) \cite{zellinger2019robust}. CMD's superiority over other methods stems from its ability to match not only means but also high-order moments. Moreover, CMD remains relatively immune to the impact of the regularization term's weight that gets added to the loss function \cite{long2017deep}. As a pragmatic distance metric, CMD quantifies the distance between probability distributions $P$ and $Q$:
\begin{equation}
\text{CMD} = \frac{1}{|b-a|} ||E(P)-E(Q)||_2+\sum_{k=2}^\infty \frac{1}{|b-a|^k} ||c_k(P)-c_k(Q)||_2
\label{eq:cmd}
\end{equation}
where $c_k$ denotes the $k$-th order moment and $a, b$ represent the joint distribution support of the distributions. In practice, only a finite number of moments are included (e.g., $k=5$). We denote the CMD measure as $D(P, Q)$. 
The objective function of EP is 
\begin{equation}
    \min_{\theta_{ep}} \mathcal{L}_{ep} = \min_{\theta_{ep}} BCE(f_{ep}(\mathbf{A}, \mathbf{X}), \mathbf{A}) + \alpha D(\mathbf{Z}_{U}, \mathbf{Z}_{\mathcal{C}_A})
    \label{eq:edge_predictor}
\end{equation}
where $BCE$ represents the Binary Cross Entropy loss function, $\alpha$ represents a hyperparameter that determines the weight of the distribution distance, $\mathbf{Z}_{U}$ and $\mathbf{Z}_{\mathcal{C}_A}$ refer to node representations on unlabeled test nodes and (pseudo-)labeled nodes, respectively. These are obtained from the teacher model in Eq.(\ref{eq:Z}). 
We refer to this module as the \textbf{EP(CMD)} module, distinguishing it from a pure \textbf{EP} module which does not involve CMD minimization.




The training process for this step is delineated from lines 3 to 6 in Algorithm \ref{alg1} found in Appendix \ref{app:algo}. 
To commence, 
a teacher GNN $f_{\theta_t}$ and an Edge Predictor (EP) $f_{ep}$ are collaboratively trained using a joint loss function across several epochs as 
\begin{align}
    \min_{\theta_t, \theta_{ep}} \mathcal{L}_{teacher} = \min_{\theta_t, \theta_{ep}} CE(f_{\theta_t}(\mathbf{A}', \mathbf{X}), \mathbf{Y}_{\mathcal{C}_A})  + \beta \mathcal{L}_{ep} 
    \label{eq:teacher}
\end{align}
where 
the label set on (pseudo-)labeled nodes, $\mathcal{C}_A$, is denoted as $\mathbf{Y}_{\mathcal{C}_A}$, and $\beta$ is a hyperparameter that regulates the weight of the EP loss. It is crucial to highlight that the teacher GNN and edge predictor are trained jointly over multiple stages. For computational efficiency, the pre-trained version is trained in the first stage. 

Sampling adjacency matrices carries a complexity of $O(|\mathcal{V}|^2)$. To enhance efficiency, we pre-select a subset of nodes likely to induce a distribution shift. This subset comprises $m$ nodes and $e$ edges, with $m \ll |\mathcal{V}|$ and $e \ll |\mathcal{E}|$.  In generating a new adjacency matrix variant, we only alter edges around these $m$ nodes and $e$ edges, preserving others, which reduces complexity to $O(m|\mathcal{V}| + e)$. We apply two selection criteria. Firstly, we pick the top-ranked test nodes with larger CMD distances to the training nodes, and reciprocally, training nodes with larger CMD distances to the test nodes. This adjustment optimizes the distributional consistency between training and test nodes in the generated graph variants while limiting node modifications. Secondly, we use the feature similarity matrix to identify unconnected edges with high similarity and connected edges with low similarity. This approach is rooted in the understanding that feature similarity can serve as a potential indicator of an edge's existence.

\subsection{Distribution Consistency Aware High-Quality Pseudo-Label Selection} 

To alleviate the sparse label issue, we will adopt pseudo-labeling to augment the labeled dataset. However, recent study show that pseudo labeling may
introduce new distribution shifts because an increasing number of nodes with low information gain, selected based on high confidence, are incorporated into the original labeled dataset, leading to the distribution progressively shifting towards the augmented dataset and overemphasis on such low-information nodes~\cite{liu2022confidence}.
This issue will be more severe when the initial labeled set and unlabeled set have a distribution shift, which is empirically verified in Section \ref{sec:visual} as biased pseudo-labeled nodes cannot benefit the model performance. 
However, existing pseudo-labeling methods commonly overlook these shifts. Though DR-GST's \cite{liu2022confidence} can alleviate the new distribution shift introduced by pseudo labels, it assumes the initial training set and test set do not have distribution shifts, which cannot be simply applied for our setting.  
To address this challenge, we propose a novel \textbf{Distribution Consistency (DC)} criterion to identify nodes capable of compensating for the distribution shift between (pseudo-)labeled training set and test set. 
It can also tackle the initial distribution shift between labeled nodes and test nodes.




Specifically, 
the DC criterion pinpoints nodes that can effectively reduce the discrepancy. We first select the top $K$ unlabeled nodes for each class that have the largest confidence score (predicted probability) as a candidate set $\mathcal{C}$, thereby filtering out noisy labels via: 
\begin{equation} 
\mathcal{C} = \{v_u| v_u \in top_j(\mathcal{V}_U, K), j \in \mathcal{Y} \}
\label{eq:candidate}
\end{equation}
where $top_j(\mathcal{V}_U, K)$ denotes the $K$ most confident nodes in unlabeled set $\mathcal{V}_U$ for the $j$-th class, as determined by the teacher GNN. The value of $K$ is expanded to $(1+\lambda) K$ in the next stage to guarantee the addition of new pseudo labels, where $\lambda$ is the expanded ratio. Subsequently, we designate a subset $\mathcal{C}_S \subset \mathcal{C}$ that has minimal CMD between (pseudo-)labeled training set and test set, i.e.,
\begin{equation}
\mathcal{C}_S = \arg\min_{\mathcal{C}' \subset \mathcal{C}} D(\mathbf{Z}_{U}, \mathbf{Z}_{\mathcal{C}' \cup \mathcal{V}_L}), \text{ s.t. } |\mathcal{C'}| = \delta
\end{equation}
where $\mathbf{Z}_{\mathcal{C}' \cup \mathcal{V}_L}$ is node representations of $\mathcal{C}' \cup \mathcal{V}_L$, $\mathbf{Z}_{U}$ is the node representations of the unlabeled test nodes, and $\delta$ is the size of $\mathcal{C}_S$. 

As less informative pseudo-labeled nodes may not yield substantial performance gain, we introduce another \textbf{Neighborhood Entropy Reduction (NER)} criterion to maximize the knowledge that selected pseudo labels can offer to the neighborhood. Specifically, NER can pinpoint nodes that offer substantial information on their neighborhood, thereby facilitating the acquisition of more informative pseudo-nodes. In the computation of the NER term, we take cues from the label propagation method \cite{zhu2002learning}. Specifically, we disperse the logits vector of the central node to its neighboring nodes, mediated by the edge weight. The NER term to be maximized quantifies the entropy reduction in the neighbors' softmax vectors following their addition. The underlying intuition is that a node is most informative when it can maximally reduce the uncertainty (i.e. entropy) of its neighborhood. 
This NER term is calculated as:
\begin{align}
    \begin{aligned}
    &NER(N(\mathcal{C}')) = \sum_{v_c \in \mathcal{C}'} \sum_{v_n \in N(v_c)} NER(v_n) \\
    &=  \sum_{v_c \in \mathcal{C}'} \sum_{v_n \in N(v_c)} H(\sigma(\mathbf{r}_{v_n})) - H(\sigma(\mathbf{r}_{v_n}+w(v_c, v_n) \cdot \mathbf{r}_{v_c}))
    \end{aligned}
\end{align}
where $\mathcal{C}'$ denotes a subset of the candidate set, $N(v_c)$ denotes the set of $v_c$'s neighbors, and $\mathbf{r}_v$ is the logits vector of a node $v_n$. $w(v_c, v_n)$ denotes the weight from node $v_c$ to node $v_n$, while $H(\cdot)$ and $\sigma(\cdot)$ are the entropy and softmax functions, respectively.

Combining the DC and NER, the criterion for selecting pseudo-labeled nodes can be written as
\begin{equation} 
\mathcal{C}_S = \arg\min_{\mathcal{C}' \subset \mathcal{C}} D(\mathbf{Z}_{U}, \mathbf{Z}_{\mathcal{C}' \cup \mathcal{V}_L}) - \gamma NER(N(\mathcal{C}')) 
 \text{ s.t. } |\mathcal{C'}| = \delta
\end{equation}
where $\gamma$ is a hyper-parameter to balance the information contributed to the neighboring nodes. The selected high-quality distribution shift aware pseudo labels will be incorporated into the labeled set as $\mathcal{C}_A = \mathcal{C}_S \cup \mathcal{V}_L$, which can be used to better train the classifier.


However, this circumstance poses a challenging question: as the search space grows exponentially with respect to $|\mathcal{C}|$, exhaustive enumeration becomes impractical. How might we identify the optimal $\mathcal{C}_S$ according to the given criteria?  A heuristic greedy search, such as recurrently selecting a node that minimizes CMD, can result in local optimality since it neglects the interdependence among nodes. Fortunately, the differentiability of the NER and CMD terms allows us to transform this issue into an optimization problem, which can be addressed via gradient descent. We introduce a selection vector $\mathbf{q} \in \mathbb{R}^{|\mathcal{C}|}$. Each element $\mathbf{q}_i \in \{0, 1\}$ indicates whether to select the $i$-th node in $\mathcal{C}$. The constraint $sum(\mathbf{q})=\delta$ ensures the selection of $\delta$ pseudo labels. Since this discrete vector is non-differentiable, we relax it to a continuous vector with $\mathbf{q}_i \in [0,1]$. When calculating $D(\mathbf{Z}_{U}, \mathbf{Z}_{\mathcal{C}' \cup \mathcal{V}_L})$, we retrieve all representations in $\mathcal{C}$ and weigh them by $\mathbf{q}$. The representations $\mathbf{Z}_{\mathcal{C}' \cup \mathcal{V}_L}$ consists of the weighted representations of $\mathcal{C}$ and representations of $\mathcal{V}_L$.  Likewise, when computing $NER(N(\mathcal{C}'))$, we calculate $\sum_{v_n \in N(v_c)} NER(v_n)$ for each node $v_c \in \mathcal{C}$ and weigh them by $\mathbf{q}$. Hence, we can formalize this as an optimization problem as 
\begin{align}
\mathbf{q} &=\arg\min_\mathbf{q} \quad D(\mathbf{Z}_{U}, \mathbf{Z}_{\mathbf{q} \star \mathcal{C} \cup \mathcal{V}_L})   - \gamma NER(N(\mathbf{q}\star \mathcal{C})) & \notag \\
&= \frac{1}{|b-a|} ||E(\mathbf{Z}_{U})-E(\mathbf{q} \star \mathcal{C} \cup \mathcal{V}_L)||_2 & \notag \\
& + \sum_{k=2}^\infty \frac{1}{|b-a|^k} ||c_k(\mathbf{Z}_{U})-c_k(\mathbf{q} \star \mathcal{C} \cup \mathcal{V}_L)||_2 & \notag \\ 
& - \gamma \sum_{v_c \in \mathcal{C}}\mathbf{q}_{ind(v_c)} \sum_{v_n \in N(v_c)} NER(v_n) , \quad & \notag \\
& \text{s.t.} \quad || \mathbf{q} ||_1 = \delta, \mathbf{q}_i \in [0, 1] &
\label{eq:optimization}
\end{align}
where $\mathbf{q}$ serves as the sole learnable parameter. The term $\mathbf{q} \star \mathcal{C}$ denotes a subset $\mathcal{C}'$. During optimization, $E(\mathbf{q} \star \mathcal{C} \cup \mathcal{V}_L)$ represents the expectation on node representations of $\mathcal{C}$ weighted by $\mathbf{q}$ and $\mathcal{V}_L$, and $ind(v_c)$ indicates the index of node $v_c$ in $\mathcal{C}$. We then formulate the loss function as follows:
\begin{align} 
    \mathcal{L}_q = D(\mathbf{Z}_{U}, \mathbf{Z}_{\mathbf{q} \star \mathcal{C} \cup \mathcal{V}_L}) - \gamma NER(N(\mathbf{q}\star \mathcal{C})) + \max(0, ||\mathbf{q}||_1 - \delta ) 
    \label{eq:loss_q}
\end{align} 
During optimizing, we project the variable $\mathbf{q}_i$ into the range $[0,1]$ by clipping. 
Upon determining $\mathbf{q}$, we can identify the top $\delta$ nodes in $\mathcal{C}$ as $\mathcal{C}_S$. This procedure is outlined in lines 8 to 11 of Algorithm \ref{alg1} found in Appendix \ref{app:algo}. We then repeat the process of training the teacher GNN and generating pseudo-labels across multiple stages until \textit{stopping condition} is reached. We establish the stopping condition based on the convergence of the CMD metrics throughout multiple stages. This iterative approach capitalizes on the utility of the multi-stage self-training framework, offering an advantage over competing methodologies.

\subsection{Student GNN Training}

In alignment with the paradigm of the traditional self-training framework \cite{li2018deeper}, we train a student GNN $f_{\theta_s}$ to fully utilize the pseudo labels generated by the teacher GNN in the final stage. To maximize the performance of the student GNN, we initialize it with the parameters of the best teacher GNN, supply it with the original graph data, and omit the edge predictor. The loss function is formulated as: 
\begin{equation}
\min_{\theta_s} \mathcal{L}_{student} = \min_{\theta_s} CE(f_{\theta_s}(\mathbf{A}, \mathbf{X}), \mathbf{Y}_{\mathcal{C}_A})
\label{eq:student}
\end{equation}
where $\mathcal{C}_A$ denotes the final augmented set. The student GNN ultimately delivers predictions for all initially unlabeled nodes. The final algorithm is detailed in Appendix \ref{app:algo}, while its complexity analysis is described in Appendix \ref{app:time}. 
\section{Experiments}
In this section, we conduct experiments on real-world datasets to evaluate the effectiveness of the proposed framework.

\begin{table*}[t]  
\centering
\caption{Node classification results (ACC$\pm$std) with biased training samples on Cora, PubMed, and Ogbn-arxiv. (bold: best)}
\vskip -1em
\small
\begin{tabular}{c|ccccc|ccc|cc}

\hline
&\multicolumn{5}{c|}{Cora}&\multicolumn{3}{c|}{PubMed}&\multicolumn{2}{c}{Ogbn-arxiv}\\
Label Rate&0.5\%&1\%&2\%&3\%&5\%&0.03\%&0.05\%&0.1\%&1\%&5\%\\
\hline
GCN&47.5$\pm$8.3&54.5$\pm$10.1&64.0$\pm$6.1&71.2$\pm$3.4&73.0$\pm$1.5 &48.6$\pm$ 4.6 &52.0$\pm$5.7&54.3$\pm$3.7 &53.2$\pm$1.3&58.7$\pm$0.3\\
\hline
ST&49.7$\pm$8.2&55.2$\pm$9.7&64.6$\pm$5.6&72.1$\pm$4.1&73.7$\pm$2.0&50.6$\pm$3.4&53.4$\pm$5.5&54.2$\pm$3.7 &54.6$\pm$2.0&58.0$\pm$0.94\\
M3S&50.0$\pm$8.5&55.9$\pm$9.8&65.3$\pm$5.6&72.9$\pm$4.3&74.6$\pm$2.2 &32.1$\pm$4.3&37.7$\pm$4.4&51.2$\pm$3.4 &54.6$\pm$2.1&59.6$\pm$1.4\\
DRGST&56.9$\pm$8.8&59.9$\pm$6.8&72.1$\pm$3.1&75.6$\pm$2.4&77.5$\pm$1.4&49.8$\pm$4.6&52.4$\pm$5.7&59.1$\pm$7.1&53.7$\pm$1.0&60.7$\pm$1.6\\
\hline
SRGNN&47.8$\pm$5.7&58.3$\pm$7.9&65.9$\pm$5.6&72.1$\pm$3.6&74.0$\pm$1.5 &50.4$\pm$2.7&53.1$\pm$5.5&56.2$\pm$6.6&55.3$\pm$1.4&58.2$\pm$0.8\\
EERM&50.8$\pm$6.6&60.1$\pm$4.2&66.6$\pm$3.1&62.1$\pm$6.5&68.9$\pm$3.7 &-&-&-&-&-\\
\hline
SRGNN$+$ST&64.8$\pm$5.7&67.7$\pm$4.8&74.2$\pm$7.3&75.5$\pm$3.6&79.6$\pm$2.7&59.7$\pm$3.8&60.7$\pm$6.7&64.7$\pm$4.6 &57.0$\pm$0.3&62.5$\pm$0.2\\
Ours&\textbf{65.0$\pm$8.3}&\textbf{68.2$\pm$7.0}&\textbf{77.4$\pm$4.5}&\textbf{79.8$\pm$4.3}&\textbf{80.0$\pm$2.7}&\textbf{62.7$\pm$2.2}&\textbf{67.8$\pm$3.1}&\textbf{69.3$\pm$6.6}&\textbf{63.0$\pm$0.7}&\textbf{64.8$\pm$0.2}\\
\hline
\end{tabular}
\label{tab:main_on_gcn_cora_bias}
\vspace{-1em}
\end{table*}

\begin{table}[!ht] 
\centering
\caption{Node classification results (ACC$\pm$std) with biased training samples on Citeseer. (bold: best)}
\vskip -1em 
\small
\begin{tabular}{c|cccc}

\hline
&\multicolumn{4}{c}{Citeseer}\\
Label Rate&0.5\%&1\%&2\%&3\%\\
\hline
GCN& 38.6 $\pm$ 3.4	 & 48.2 $\pm$ 5.1	 & 54.6 $\pm$ 1.6	 & 59.3 $\pm$ 0.7\\
\hline
ST& 38.8 $\pm$ 4.5	& 48.8 $\pm$ 5.8	& 57.6 $\pm$ 1.5	& 60.8 $\pm$ 0.9 \\
M3S& 38.9 $\pm$ 4.6	& 49.0 $\pm$ 5.9	& 57.9 $\pm$ 1.9	& 61.4 $\pm$ 1.1 \\
DRGST& 45.7 $\pm$ 6.4	& 49.7 $\pm$ 3.6	& 56.7 $\pm$ 2.2	& 60.6 $\pm$ 2.1 \\
\hline
SRGNN& 38.4 $\pm$ 7.5	& 48.4 $\pm$ 6.6 	& 59.2 $\pm$ 1.5	& 61.5 $\pm$ 1.5\\
EERM& 40.0 $\pm$ 7.5 	& 43.6 $\pm$ 7.1 	& 49.7 $\pm$ 9.1 	& 46.4 $\pm$ 9.8\\
\hline
SRGNN$+$ST& 53.6 $\pm$ 8.2 	  & 58.1 $\pm$ 6.0 	  & 62.8 $\pm$ 1.7 	  & 65.1 $\pm$ 3.3 \\
Ours& \textbf{58.1 $\pm$ 6.4} 	  & \textbf{62.2 $\pm$ 6.0 }	  & \textbf{65.9 $\pm$ 2.4} 	  & \textbf{67.7 $\pm$ 4.8} \\
\hline
\end{tabular}
\label{tab:main_on_gcn_citeseer_bias}
\vspace{-1em}
\end{table}

\begin{table}[!ht]
\centering
\caption{Ablation study results (ACC$\pm$std) for different components.}
\vspace{-1em}
\small
\begin{tabular}{c|c|c}
\hline
Methods & Cora & Citeseer \\
\hline
DCGST (Full)	 & 77.4$\pm$4.5 & 65.9$\pm$2.4 \\
DCGST: DC(NER)$+$EP	 & 76.9$\pm$5.3 & 65.8$\pm$2.5 \\
DCGST: DC$+$EP(CMD)	 & 75.1$\pm$4.6 & 64.8$\pm$3.1 \\
 DCGST: DC(NER) &  73.3$\pm$6.1 & 63.2$\pm$3.9 \\ 
\hline
\end{tabular}
\label{tab:ablation1}
\end{table}

\subsection{Experimental Setting}

\subsubsection{Evaluation Protocol}
For a thorough assessment of the proposed framework, we adopt four widely used graph datasets, i.e., Cora, Citeseer, PubMed \cite{sen2008collective}, and Ogbn-arxiv \cite{hu2020open}. In order to fully assess our framework, we vary label rates $|\mathbf{Y}_L|/|\mathcal{V}|$ 
from extreme few-shot scenarios to standard semi-supervised settings. 
To explore DC-GST's capability with a biased training set, we adopt a biased sampler, PPR-S \cite{zhu2021shift}, which introduces greater distribution shifts in the training set for most experiments. 
Performance is reported using accuracy (ACC) as the evaluation metric. For equitable comparison, the mean values and standard deviations from 10 independent runs are provided for all methods and scenarios. More experimental details can be found in Appendix \ref{app:dataset}.

\subsubsection{Baselines}
In current practice, there exists no fully consistent strategy to tackle the distribution shift issue during self-training. Thus, we juxtapose our proposed approach with five baseline methods, segmented into two categories: self-training methods (G1) and methods addressing distribution shift (G2). For G1, we utilize three graph self-training methods, namely, ST \cite{li2018deeper}, M3S \cite{sun2020multi}, and DR-GST \cite{liu2022confidence}. Of these, DR-GST most closely aligns with our problem setting, as it claims to recover the distribution after pseudo-labeling the same as the distribution on the training set. For G2, we adopt two prevalent approaches, SR-GNN \cite{zhu2021shift} and EERM \cite{wu2022handling}, designed to alleviate the distribution shift between training and testing sets. To ensure an equitable comparison, these methods are assessed on the identical GNN backbones such as GCN \cite{kipf2017semisupervised}, GAT \cite{velivckovicgraph}, and GSAGE \cite{hamilton2017inductive}, and with consistent dataset splitting. 
The implementation of DC-GST can be found in Appendix \ref{app:hyperparameter}.

\subsection{Node Classification with Sparse Labels}

\subsubsection{Performance on Graph with Distribution Shift}
In this subsection, we present a comparative analysis of our proposed DC-GST framework and baseline models utilizing a GCN backbone, trained on biased samples across four major citation benchmarks as elaborated in Table \ref{tab:main_on_gcn_cora_bias} and Table \ref{tab:main_on_gcn_citeseer_bias}. ``-'' represents the OOM result. These benchmarks include three small graphs and one large graph. From the two tables, we can observe: (i) Self-training based methods (G1) exceed the performance of the backbone model, primarily due to the advantages provided by pseudo labels. (ii) Among G1 models, DR-GST outperforms others due to its efficacy in averting the distribution shift introduced by pseudo-labeling, relative to ST and M3S. (iii) Methods addressing distribution shift (G2) enhance performance in comparison to the GCN model when label rates are high, but display limited advancement in a few-shot setting. (iv) The integration of SR-GNN with ST to mitigate distribution shifts during self-training results in significant performance advancement, highlighting the effectiveness of managing both distribution shifts and self-training. (v) Our method, with its carefully architected framework considering neighbor information, is capable of further performance improvements. More experimental results can be found in Appendix \ref{app:clean} and Appendix \ref{app:gen}. 

\subsection{Ablation Study}
We undertook ablation studies on the Cora and Citeseer datasets, employing the GCN backbone, to dissect the individual contributions of different components within the DC-GST. The summarized results are presented in Table \ref{tab:ablation1}. ``DCGST (Full)'', the third row in the table, represents our original experimental findings. The \textbf{CMD on EP} module's influence is assessed by removing the CMD term in Eq. \ref{eq:edge_predictor}, denoted as ``DCGST: DC(NER)$+$EP''. The effect of the \textbf{NER} module is appraised by removing the NER term in Eq. \ref{eq:optimization} and Eq. \ref{eq:loss_q}, designated as ``DCGST: DC$+$EP(CMD)''. The \textbf{EP} module's significance is verified by omitting the BCE loss in Eq. \ref{eq:teacher}, referred to as ``DCGST: DC(NER)''. From the table, we observe that: 
\begin{itemize}
\setlength{\itemsep}{0pt}
\setlength{\parskip}{-1pt}
\setlength{\parsep}{0pt}
\setlength{\leftmargin}{0pt}
    \item Compared with DCGST: DC(NER), DCGST: DC(NER)+EP exhibits superior performance across both tasks. This is because the teacher GNN offers more reliable pseudo-nodes by training it on diverse graph variants provided by the edge predictor.
    \item When excluding the NER module from DCGST results in performance degradation of 2.3\% and 1.1\% in Cora and Citeseer, respectively. This highlights the crucial role of the informativeness measurement, which effectively disseminates valuable information to neighboring nodes, in generating high-quality pseudo-labels.
    \item For DCGST: DC(NER)+EP, the absence of CMD minimization in the edge predictor leads to underperformance in accuracy when compared with DCGST (Full). This indicates that the generation of distributionally consistent graph variants can enhance the teacher model's generalization, thereby improving the overall self-training performance. 
\end{itemize}

\subsection{Analysis of Hyper-parameter Impact}
\begin{figure}[!t]
    \centering
    \includegraphics[width=0.8\linewidth]{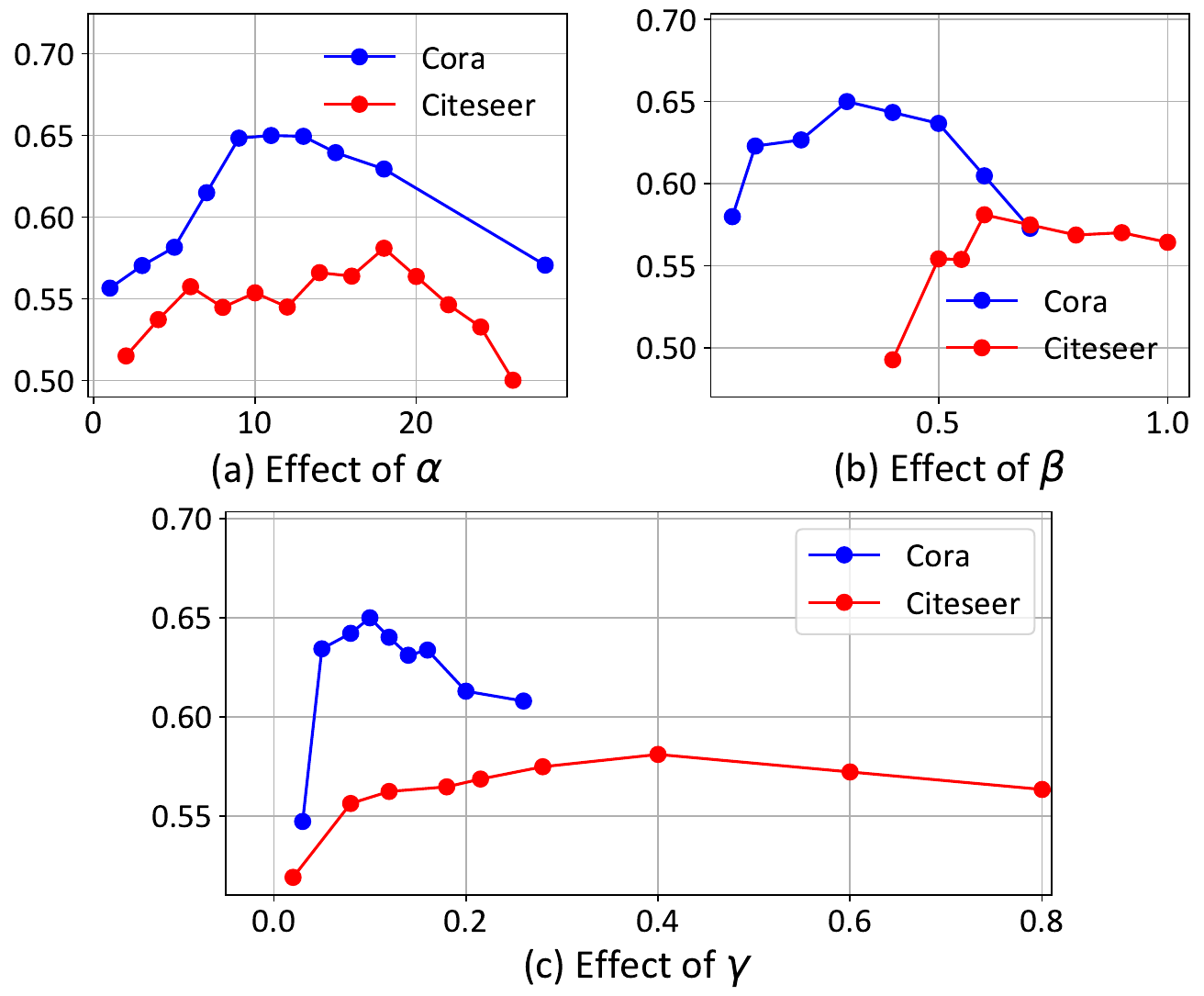}
    \vskip -1em
    \caption{Sensitivity analysis of hyper-parameters w.r.t. $\alpha$, $\beta$, and $\gamma$ on Cora and Citeseer with 0.5\% label rate. } 
    \label{fig:sensitivity}
    \vskip -1em
\end{figure}

We assess the impact of the DC-GST on its hyper-parameters, $\alpha$, $\beta$, and $\gamma$, which regulate the \textbf{CMD on EP}, \textbf{EP}, and \textbf{NER} modules respectively. This analysis is conducted on the Cora and Citeseer datasets, utilizing the GCN backbone with a 2\% label rate. 
We choose a range centered around the optimal hyper-parameter value, so we vary the values of $\alpha$ as $\{1, 3, \dots, 18, 28\}$ and $\{ 2,4,\dots,26 \}$ for Cora and Citeseer, respectively. $\beta$ is changed from $\{0.05, 0.1, \dots,0.7\}$ and $\{0.4, 0.5, \dots, 1\}$. And $\gamma$ is varied as $\{ 0.03, 0.05,$ $ \dots, 0.26 \}$ and $\{0.02, 0.08, \dots, 0.8\}$. 
As illustrated in Figure \ref{fig:sensitivity}, we observe that 
(i) While the model is fairly robust to minor deviations in the optimal values of three key hyper-parameters, extreme values negatively impact performance. Empirical evidence suggests that larger hyper-parameter values are chosen for sizable datasets, limited training instances, and an extensive feature set. 
(ii) A rightward shift in peak values of three hyper-parameters for Citeseer in comparison to Cora is observed, likely due to Citeseer's denser feature and sparser topology, making it more likely to overfit. Thus, Citeseer demands superior generalization and a greater number of representative pseudo-nodes, corresponding to increased values of $\alpha$, $\beta$, and $\gamma$.
\subsection{Benefits of Reducing Distribution Shift}
\label{sec:visual}

\begin{figure} [!htbp]
    \centering
    \includegraphics[width=0.96\linewidth]{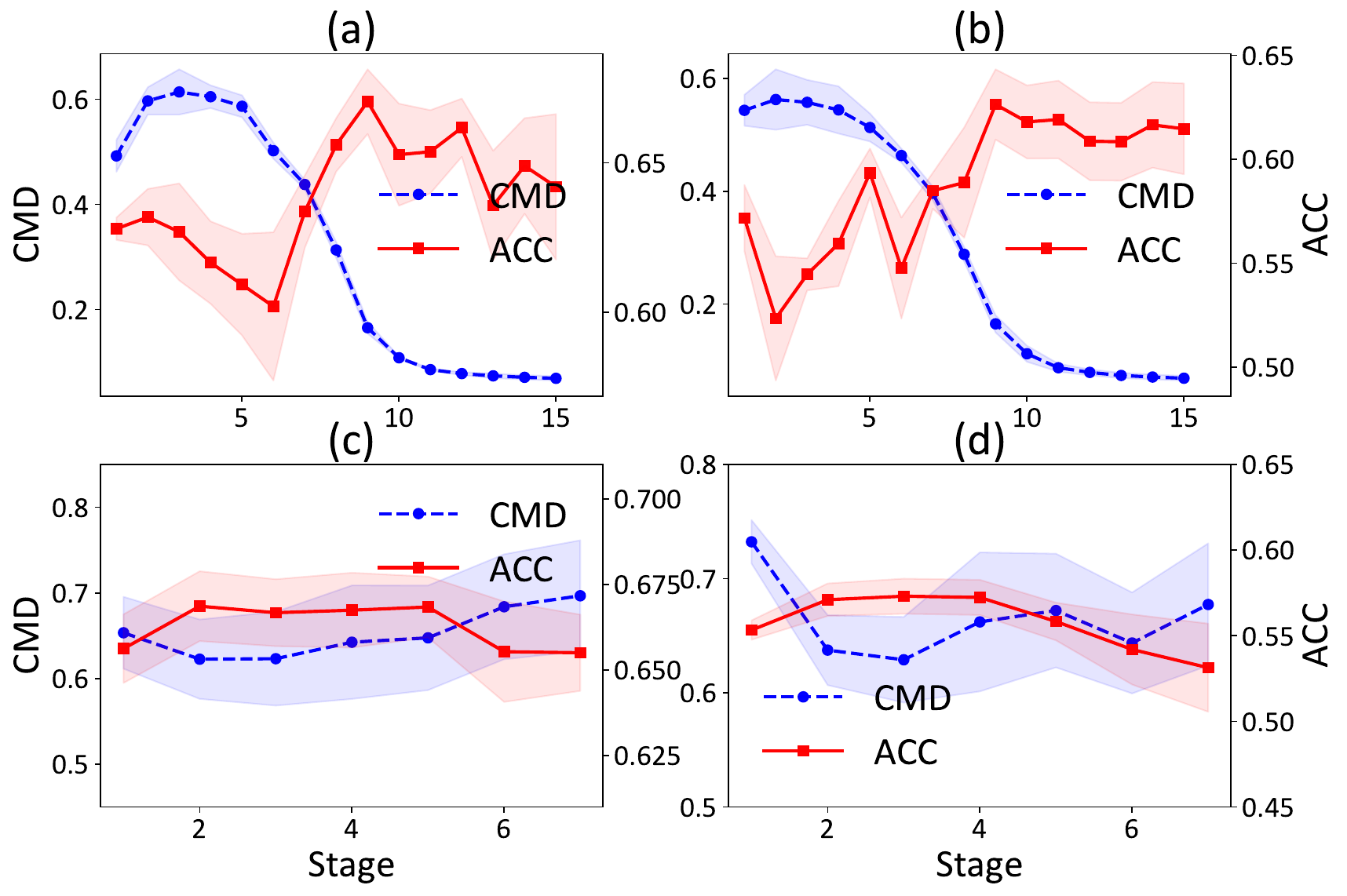}
    \vskip -1em
    \caption{Visual Analysis of CMD and Accuracy on Citeseer with a 2\% Label Rate. Sub-figures (a) and (b) illustrate CMD and ACC for DC-GST on random and biased training samples respectively, while (c) and (d) display M3S's CMD and ACC under the same conditions. The shadowed area refers to the variance of the results of the 10 runs. }
    \label{fig:visual}
    \vskip -1.5em
\end{figure}
To emphasize the advantages of explicit distribution shift reduction, we present a visual analysis comparing the ACC as an indicator of model performance and the CMD as a measure of distribution shift over multiple stages. We conduct this analysis for both our proposed method and the M3S method, which has been claimed to be effective at multiple stages of the training process. As demonstrated in Figure \ref{fig:visual}, our observations reveal that: 
(i) CMD and ACC exhibit an inverse correlation, implying that mitigating distribution shifts can enhance model performance. In later stages, CMD converges, fulfilling the stopping condition, while ACC remains relatively steady, albeit less so than CMD. 
(ii) While the M3S method is successful in reducing distribution shifts during the initial stages of training, leading to a performance gain, it fails to maintain this trend as training progresses. In contrast, our method is designed to actively and continuously minimize distribution shifts, facilitating sustained performance improvement across multiple stages of the training process. 

\section{Conclusion}
In this study, we propose DC-GST, a self-training framework based on distribution consistency, devised to counter the issue of distribution shift during self-training. Current graph self-training methodologies often rely on the assumption that independent and identically distributed (IID) samples are available for training, which neglects the inherent distribution shift within the training set, consequently inhibiting the performance of multi-stage self-training. Our solution is the distribution consistency selection criterion, which compensates for the distribution shift, coupled with a neighborhood entropy increase criterion for identifying informative pseudo-nodes. 
The effectiveness of DC-GST is well-documented through our extensive experiments. 

\begin{acks}
This material is based upon work supported by or in part by the National Science Foundation (NSF) under grant number IIS-1909702, Army Research Office (ARO) under grant number W911NF-21-1-0198, Department of Homeland Security (DHS) CINA under grant number E205949D, and Cisco Faculty Research Award. 
\end{acks}

\bibliographystyle{ACM-Reference-Format}
\balance
\bibliography{sample-base}

\newpage
\appendix
\section{Supplement}

\subsection{Algorithm}
\label{app:algo}
The algorithm of the DC-GST method is presented in Algorithm \ref{alg1}.

\begin{algorithm}[t]
\caption{Distribution Consistency-based Graph Self-training.}
\label{alg1}
\begin{algorithmic}[1]
\Procedure{PseudoLabeling}{$\mathcal{G}, \mathbf{A}, \mathbf{X}, \mathcal{V}_L, \mathcal{V}_U$}\Comment{Graph $\mathcal{G}=(\mathcal{V},\mathcal{E}, \mathbf{X})$, adjacency matrix $\mathbf{A} \in \mathcal{R}^{|\mathcal{V}|\times |\mathcal{V}|}$, features $\mathbf{X}$, labeled nodes $\mathcal{V}_L$, unlabeled nodes $\mathcal{V}_U$}
\State Partition labeled nodes into training and validation sets
\State Pretrain a GNN on the training set
\While{stopping condition not met} \Comment{Execute for each stage} 
\State \textit{\# Step 1: Training a Teacher Model}
\State Jointly train a teacher GNN and an edge predictor on training data using Eq. \ref{eq:teacher}
\State \textit{\# Step 2: Selecting Pseudo-Labeled Nodes}
\State Select $K$ candidate nodes using Eq. \ref{eq:candidate}
\State Optimize $\mathbf{q}$ using Eq. \ref{eq:optimization} and Eq. \ref{eq:loss_q}
\State Select top $\delta$ nodes in $\mathbf{q}$
\State Update training data and increment $K$ by $(\lambda+1) K$
\EndWhile\label{euclidendwhile}
\State \textit{\# Step 3: Training a Student Model}
\State Train a student GNN on the updated training data and make predictions for $\mathcal{V}_U$ using Eq. \ref{eq:student}
\State \textbf{return} the student GNN and its predictions for $\mathcal{V}_U$
\EndProcedure
\end{algorithmic}
\end{algorithm}

\subsection{Time Complexity Analysis}
\label{app:time}
Assuming a standard GCN \cite{kipf2017semisupervised} model is applied in self-training, the training time complexity for a GCN with $L$ layers amounts to $\mathcal{O}(L|\mathcal{E}|F + L|\mathcal{V}|F^2)$, where $F$ denotes the dimension of the learned representations. Within the DC-GST framework, the time complexity of the edge predictor aligns with that of a GCN. Additionally, the time complexity of the pseudo-label selection procedure is $\mathcal{O}(k|\mathcal{V}| |\mathcal{Y}| + d|\mathcal{C}||\mathcal{Y}|)$, where $k=5$ signifies the number of moments in CMD, $d$ represents the average degree, and $|\mathcal{Y}|$ denotes the size of the label set. The training process for a student GCN equates to training a standard GCN. It is crucial to note that the complexity of both the edge predictor and selection step is less than or equivalent to that of the GCN, thereby rendering the overall time complexity in DC-GST acceptable in comparison with the training of the standard GCN classifier within self-training. 

\subsection{More Experimental Details}

\subsubsection{Datasets and Evaluation Metrics} 
\label{app:dataset}
For a thorough assessment of the proposed framework, we adopt four widely used graph datasets, i.e., Cora, Citeseer, PubMed \cite{sen2008collective}, and Ogbn-arxiv \cite{hu2020open}. Detailed statistics for these datasets can be found in Table \ref{tab:dataset_statistics}. In order to fully assess our framework, we vary label rates $|\mathbf{Y}_L|/|\mathcal{V}|$ 
from extreme few-shot scenarios to standard semi-supervised settings. For instance, a 0.5\% label rate equates to 2 labels per class on Cora, and a 5\% rate corresponds to 19 labels per class, which is roughly equivalent to the standard split.
Label rates of 0.5\%, 1\%, 2\%, 3\%, and 5\% are tested on Cora and CiteSeer, while rates of 0.03\%, 0.05\%, 0.1\%, and 0.3\% are tested on PubMed, and 1\% and 5\% on Ogbn-arxiv. These label rates are chosen to facilitate straightforward comparison with baseline studies. 
To explore DC-GST's capability with a biased training set, we adopt a biased sampler, PPR-S \cite{zhu2021shift}, which introduces greater distribution shifts in the training set for most experiments. Performance on an unbiased training set, i.e., randomly sampled, will also be reported. Across all datasets and label rates, 0.5\% of nodes are randomly designated as validation sets, with the remainder constituting the test nodes (also referred to as unlabeled nodes).
Performance is reported using accuracy (ACC) as the evaluation metric. For equitable comparison, the mean values and standard deviations from 10 independent runs are provided for all methods and scenarios.
Note that the selected graph datasets exhibit homophily; datasets displaying heterophily are left for future work.


\begin{table}[!t]
    \centering
    \caption{Dateset statistics. Label Rate refers to $|\mathbf{Y}_L|/|\mathcal{V}|$ in the standard split.}     \label{tab:dataset_statistics}
    \vspace{-1em}
    \small
    \begin{tabular}{cccccc}
        \hline 
         Dateset & Nodes & Edges & Classes & Features  & Label Rate \\
         \hline
         Cora & 2708 & 5429 & 7 & 1433 & 5.2\% \\
         Citeseer & 3327 & 4732 & 6 & 3703 & 3.6\% \\
         PubMed & 19717 & 44338 & 3 & 500 & 0.3\% \\ 
         Ogbn-arxiv & 169343 & 1166243 & 40 & 128 & 53.7\% \\
        \hline
    \end{tabular}
\end{table}

\subsubsection{Hyper-parameters} 
\label{app:hyperparameter}
The pivotal hyperparameters within the proposed framework—$\alpha$, $\beta$, and $\gamma$—modulate the influence of the EP, CMD on EP, and NER modules, respectively. These hyperparameters are tuned using the Bayesian search method facilitated by W\&B \cite{wandb}. Specifically, $\alpha=8$, $\beta=0.3$, and $\gamma=0.1$ are established for Cora; $\alpha=18$, $\beta=0.6$, and $\gamma=0.4$ for Citeseer; $\alpha=13$, $\beta=1.5$, and $\gamma=0.7$ for PubMed; and $\alpha=26$, $\beta=0.7$, and $\gamma=0.6$ for Ogbn-arxiv. These values are determined based on the GCN backbone. Furthermore, the expanded ratio $\lambda$ is set to 0.5, the initial number of most confident nodes $K$ is equal to the number of grounded-truth nodes per class, the number of top nodes $\delta$ is $K/2$, and the learning rate is assigned as 0.01 for Cora and Citeseer, and 0.001 for PubMed and Ogbn-arxiv, with L2 regularization set to 5e-4. The model is trained for multiple stages using the Adam optimizer \cite{kingma2014adam}. The stopping condition is met when the CMD fails to decrease over five stages. All hyperparameters of baselines are from those suggested in their original papers. 

\subsection{More Experimental results}
\subsubsection{Performance on Randomly Sampled Training Sets Without Crafted Biases} 
\label{app:clean}
As shown in Table \ref{tab:exp_clean}, we conduct experiments on randomly sampled training sets. Through the comparison of results between Table \ref{tab:main_on_gcn_citeseer_bias} and Table \ref{tab:exp_clean}, we can see the bias within the training set lead to significant performance degradation especially when the label rate is low. Moreover, in Table \ref{tab:exp_clean}, the primary distribution shift arises from the sequential addition of high-confidence pseudo-nodes to the original labeled dataset, leading to an overemphasis on these easy nodes, as delineated in DR-GST \cite{liu2022confidence}. The results reveal that our method can compete with or even outperform DR-GST, demonstrating our effectiveness in mitigating the distributional shifts caused by pseudo-labeling.

\begin{table}[!ht]
    \caption{Node classification results (ACC$\pm$std) with randomly sampled training samples on Citeseer. (bold: best)}
    \label{tab:exp_clean}
    \vskip -1em
    \small
    \centering
    \begin{tabular}{c|cccc}
         \hline					
&\multicolumn{4}{c}{Citeseer}\\	
 Label Rate  & 0.5\%  & 1\%  & 2\%  & 3\%  \\
\hline     
GCN & 47.2 $\pm$ 8.6  & 54.5 $\pm$ 4.8  & 65.0 $\pm$ 2.0  & 66.7 $\pm$ 1.4 \\    
\hline 
ST & 49.1 $\pm$ 5.0  & 55.8 $\pm$ 4.6  & 66.7 $\pm$ 1.8  & 68.9 $\pm$ 1.6 \\
M3S & 50.8 $\pm$ 5.0  & 55.6 $\pm$ 6.1  & 67.7 $\pm$ 1.6  & 69.4 $\pm$ 1.3 \\
DRGST & 53.3 $\pm$ 7.6  & 61.2 $\pm$ 3.9  & 67.6 $\pm$ 1.8  & 70.3 $\pm$ 1.3 \\
\hline                 
SRGNN & 49.3 $\pm$ 4.5 & 55.8 $\pm$ 4.1    & 65.9 $\pm$ 2.5    & 67.8 $\pm$ 1.2 \\
EERM & 48.9 $\pm$ 5.8 & 55.1 $\pm$ 4.6 & 58.5 $\pm$ 5.6 & 63.0 $\pm$ 4.3 \\
\hline                 
SRGNN$+$ST & 57.3 $\pm$ 6.9 & 61.3 $\pm$ 5.3    & 66.9 $\pm$ 2.7    & 68.7 $\pm$ 3.0 \\
Ours & \textbf{61.2 $\pm$ 5.6} & \textbf{65.3 $\pm$ 3.6 }   & \textbf{68.3 $\pm$ 1.8}    & \textbf{70.4 $\pm$ 2.9} \\
\hline 				
\end{tabular}
\end{table}

\begin{figure}[!t]
    \centering
    \includegraphics[width=\linewidth]{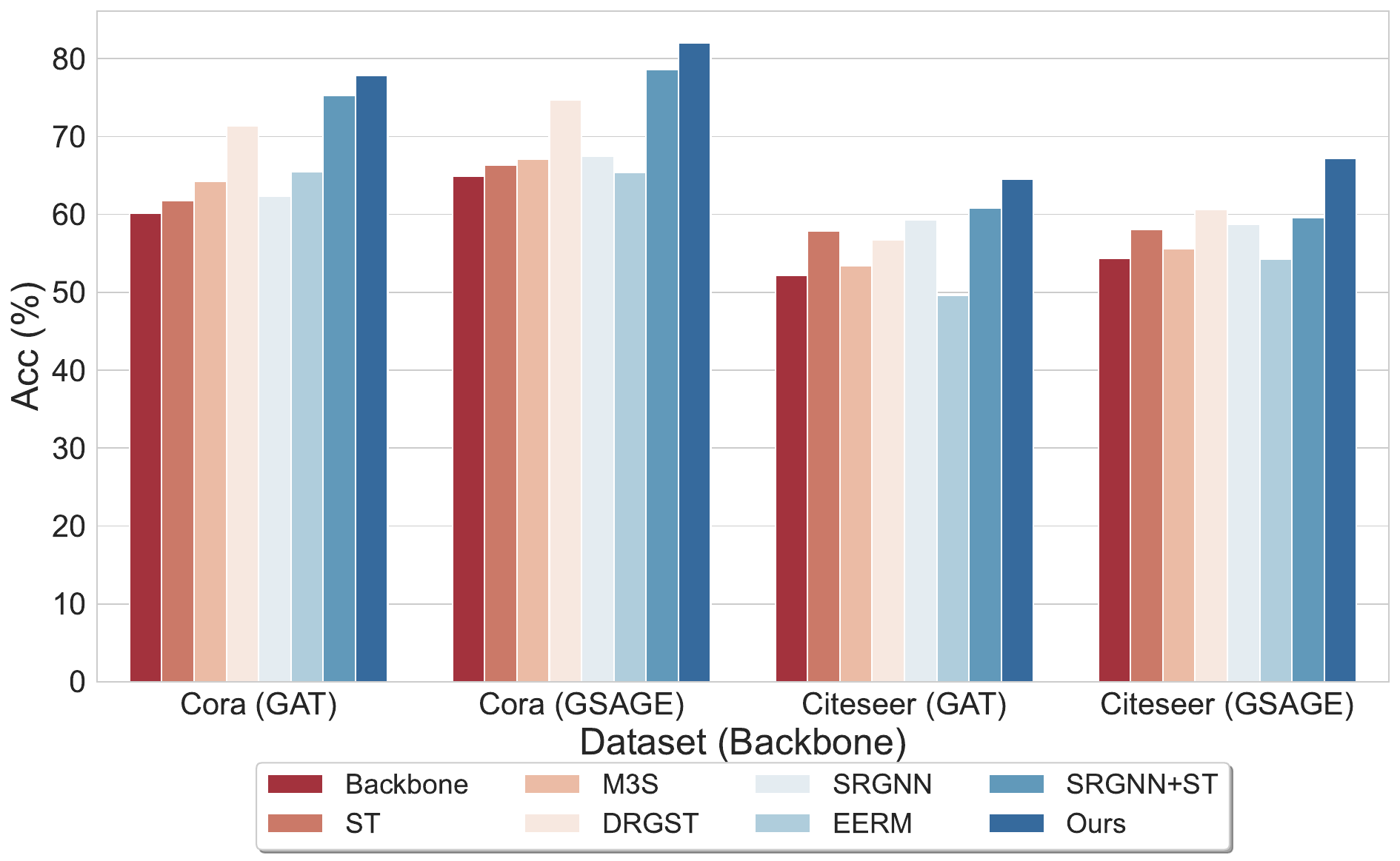}
    \vskip -1em
    \caption{Generalizability with GAT and GSAGE backbones on biased training samples.} 
    \label{fig:gat_gsage}
\end{figure}

\subsubsection{Generalizability across Different Backbones}
\label{app:gen}
To verify our self-training framework's model-agnosticism, we execute experiments using diverse backbones such as GAT and GSAGE on biased training samples with a 2\% label rate. Since we have similar observations on other datasets, we only report results on Cora and Citeseer in Figure \ref{fig:gat_gsage}. From the figure, we observe that: These experiments exhibit trends parallel to those seen in GCN, attesting to our framework's broad generalizability.

\end{document}